# Enhanced Traffic Flow Prediction with Multi-Segment Fusion Tensor Graph Convolutional Networks

Wei Zhang and Peng Tang

*Abstract*—Accurate traffic Flow Prediction can assist in traffic management, route planning, and congestion mitigation, which holds significant importance in enhancing the efficiency and reliability of intelligent transportation systems (ITS). However, existing traffic flow prediction models suffer from limitations in capturing the complex spatial-temporal dependencies within traffic networks. In order to address this issue, this study proposes a multi-segment fusion tensor graph convolutional network (MS-FTGCN) for traffic flow prediction with the following three-fold ideas: a) building a unified spatial-temporal graph convolutional framework based on Tensor M-product, which capture the spatial-temporal patterns simultaneously; b) incorporating hourly, daily, and weekly components to model multi temporal properties of traffic flows, respectively; c) fusing the outputs of the three components by attention mechanism to obtain the final traffic flow prediction results. The results of experiments conducted on two traffic flow datasets demonstrate that the proposed MS-FTGCN outperforms the state-of-the-art models.

## I. Introduction

In the past few years, spurred by the rapid advancement of intelligent transportation systems [1] and the widespread availability of diverse data sources [2] such as GPS trajectories, traffic cameras, and mobile applications [3], there has been an increasing demand for advanced traffic prediction models that can effectively utilize these data for accurate predictions. At the same time, real-time changes in dynamic transportation networks [4,5] are affected by many factors, such as time, season, and weather. Therefore, prediction models for dynamic traffic networks need to take this cyclical change into account. It is a very practical topic to make good use of existing data and real-time information to predict future traffic status to establish a more accurate and reliable prediction model. By means of a comprehensive examination of traffic data and the application of modelling techniques [6], the periodic change patterns of dynamic traffic network can be more accurately discerned, thereby enhancing the precision and practicality of predictions.

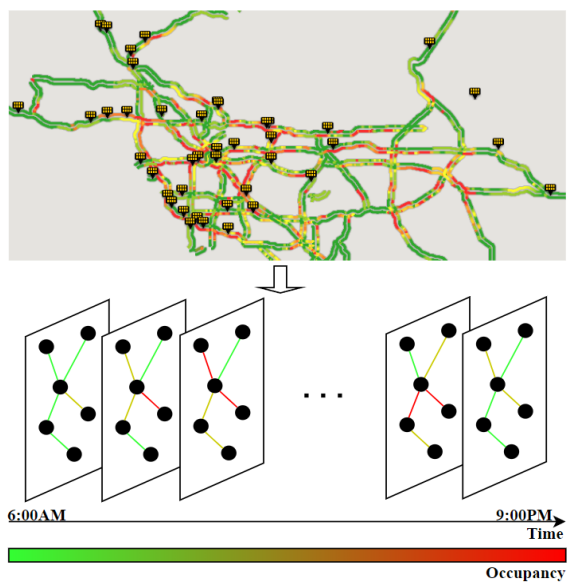

Fig. 1. The changes of traffic flow over time

The traffic data detectors shown in Fig. 1 can be viewed as nodes in the network, while roads correspond to edges in the network. This formulation establishes a link between the structure of dynamic transportation networks and concepts from graph theory [7], further emphasizing both the dynamic nature and complexity of the latter. It can be clearly seen from Fig. 1 that the

➢ W. Zhang is with the School of Computer Science and Technology, Chongqing University of Posts and Telecommunications, Chongqing 400065, China (e-mail: zhangweihed@outlook.com).
➢ P. Tang is with the College of Computer and Information Science, Southwest University, Chongqing 400715, China (e-mail: tangpengcn@swu.edu.cn)

traffic flow of a certain location or road is not independent, but affected by surrounding nodes or edges. In other words, their states are dynamically related [8,9]. The periodicity of traffic flow is evident within a very certain time range, with traffic volume, speed and congestion showing repeated patterns of change. This periodicity can be observed based on different time scales, such as hours, days, weeks, or seasons. In cities, the periodicity of traffic conditions is usually related to daily life and work habits. Cyclical variations in traffic flow and congestion can be determined through traffic data analysis and historical records. Additionally, specific events or holidays may also impact traffic conditions. For instance, the commencement and conclusion of academic institutions, high-traffic periods during holidays, and major sporting or cultural events can result in significant alterations in traffic patterns. To summarize, extracting spatial-temporal correlations from intricate and fluctuating traffic data [10-12] and producing precise traffic forecasts is a demanding subject.

The traditional traffic flow prediction model [13] based on time series analysis utilizes the time series characteristics of historical traffic data [4,14-16] for trend analysis and periodic prediction, such as support vector regression (SVR) [17], Kalman filtering [18-20], ARIMA [21], and historical average model (HAM) [22], which have been widely studied by researchers. While these models excel at modelling traffic states, they are limited in their ability to deal with large, nonlinear, and complex spatial-temporal traffic data. Recently, researchers have focused on graph convolutional networks [23], as they offer significant advantages in handling nonlinear data [24-27]. Zhao *et al*. [28] proposed a Time Graph Convolutional Network (T-GCN) that utilizes GCN [29] to capture the spatial topology of a graph and obtain its spatial correlation; In contrast, the Gated Recurrent Unit (GRU) model is utilized to capture the dynamic changes of node attributes and to obtain the temporal dependence of node attributes. Guo *et al*. [30] introduced an attention mechanism tailored for spatial-temporal data, termed the Attention-based Spatiotemporal Graph Convolutional Network (ASTGCN). This innovative approach employs spatial attention to model intricate spatial dependencies across diverse locations and temporal attention to capture evolving temporal correlations over different time intervals. Zheng *et al*. [31] introduced the Graph Multi Attention Network (GMAN) for traffic flow prediction. This network has an encoder-decoder structure, where a transition attention layer is added between the encoder and decoder to convert the encoded traffic features to the decoder. Integrating graph structure and temporal information into multi attention mechanisms through spatial-temporal embedding. Song *et al*. [32] proposed the Traffic Flow Prediction via Spatial Temporal Graph Neural Network Forecasting (STSGCN) to effectively capture heterogeneity in spatial-temporal networks, which can directly extract local spatial-temporal correlations instead of using multiple module combinations for extraction. Yu *et al*. [33] proposed an innovative approach for constructing data-driven graphs [34], which leverages graph adjacency learning. Exploring dependencies between time series using graph attention mechanism and embedding sensor correlations into potential attention space [35-37] to determine the correlation of any possible sensor pairs used for traffic map construction. Wang *et al*. [38] proposed a graph based spatial-temporal autoencoder for predicting spatial-temporal traffic speeds with missing values. Then they employed graph convolutional layers incorporating adaptive adjacency matrices to model spatial dependencies, complemented by gated recurrent units for temporal learning. Chen *et al*. [39] proposed A Flow Feedback Traffic Prediction Based on Visual Quantified Features (V-STF), The integration of visual methods intend to enhance the accuracy of predictions during non-periodic peak hours. The most crucial aspect of traffic flow prediction is to capture its spatial-temporal correlation from a complex road network [40]. This study proposes a method based on the combination of tensor M-product [41,42] and GCN [43,44] for traffic flow prediction. The main contributions of this paper are summarized as follows:

- We design a new model to capture the spatial-temporal dependencies of traffic data. Specifically, we use the TM-GCN [45] method to embed spatial-temporal information in the dynamic graph, and then fuse the features of multiple segments to obtain the final prediction result.
- The validation results on actual datasets show that our model performs excellently in predictive performance, significantly exceeding the existing benchmark models. This further validates the efficacy and viability of our proposed approach. Compared with baseline models, our model demonstrates superior accuracy and stability.

## II. PRELIMINARIES

### A. Traffic network

In this paper, the traffic network can be defined as $G = (V, E)$, where $V = \{v_1, v_2..., v_{N-1}, v_N\}$ is the set of nodes, which refers to detectors in traffic network. $N$ is the number of nodes. $E = \{e_{ij} \mid v_i, v_j \in V\}$ is the set of edges [46-50], representing the distance between detectors. $X \in \mathbb{R}^{N \times F}$ represents the characteristics of $N$ nodes. This paper aims to identify the function $f$ to predict the future features over $S$ time steps, utilizing the traffic features observed during $T$ historical time steps:

$$f(G\ ;\ [X_{t-T+1},...,X_t]) \rightarrow [\bar{X}_{t+1},...,\bar{X}_{t+S}]. \tag{1}$$

### B. Framework of Tensor M-Product

Here, we introduce the framework of tensor M-product [45]. A tensor [51-54,56] refers to a multidimensional array of real numbers, represented in bold capital letters in text, e.g. **X**. The matrix is represented by italicized capital letters, e.g. $M$. The subscript of a tensor represents the dimensional situation of the tensor, e.g. $\mathbf{X}_{ijt}$ The colon represents all elements of the tensor in that dimension, e.g. $\mathbf{X}_{::k}$. The number $k$ on the time dimension of tensor **X** represents the $k$-th front slice of **X**.

Tensor M-product is a definition of the product of two tensors [55]. It is worth noting that the M-product of two tensors with the same dimensions remains unchanged [57-59]. The following definitions provide a detailed description of tensor M-product.

***Definition 1***: (M-Transform) $M \in \mathbb{R}^{T \times T}$ represent a mixed matrix, and the M-Transform applied to tensor $\mathbf{A} \in \mathbb{R}^{N \times N \times T}$ can be defined as:

$$(\mathbf{A} \times_3 M)_{ijt} = \sum_{k=1}^{T} M_{tk} \mathbf{A}_{ijk}. \tag{2}$$

It can be considered that $\mathbf{A} \times_3 M$ is situated within the transformed space. Additionally, ***Definition 1*** can also be represented in form $(\mathbf{A} \times_3 M)_{ijt} = \text{fold}(M \times \text{unfold}(\mathbf{A}))$. The method used is to unfold the tensor $\mathbf{A}$ and perform an M-transform, subsequently stacking it into a new tensor.

***Definition 2***: (Face-wise Product) Consider $\mathbf{A} \in \mathbb{R}^{N \times N \times T}$ and $\mathbf{B} \in \mathbb{R}^{N \times F \times T}$ as two tensors, where their face-wise product can be defined as:

$$(\mathbf{A} \ominus \mathbf{B})_{::t} = \mathbf{A}_{::t} \mathbf{B}_{::t}. \tag{3}$$

***Definition 3***: (M-Product) From ***Definition 1*** and ***Definition 2***, we can summarize the final definition of M-product, where $\mathbf{A} \odot \mathbf{B} \in \mathbb{R}^{N \times F \times T}$.

$$\mathbf{A} \odot \mathbf{B} = ((\mathbf{A} \times_3 M) \ominus (\mathbf{B} \times_3 M)) \times_3 M^{-1}. \tag{4}$$

In the original expression of M-product, the matrix $M$ can be computed using fast Fourier transform, given that $M$ is a discrete Fourier transform matrix. This framework was subsequently extended to any invertible matrix $M$. The complete graphical calculation process of M-product is illustrated in Fig. 2 where $\mathbf{X}$ and $\mathbf{A}$ are feature tensors and adjacency tensors respectively.

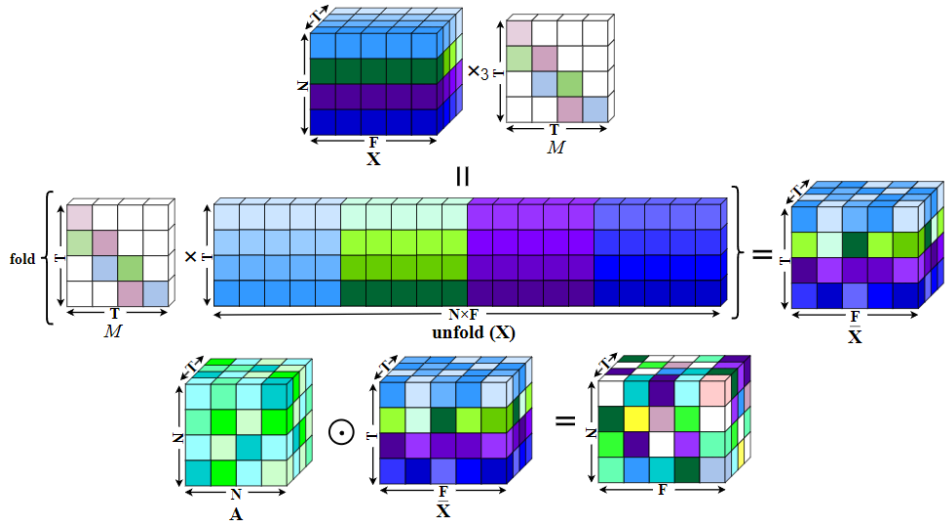

Fig. 2. The calculation process of tensor M-Product

III. THE PROPOSED MODEL

A. Tensor M-Product Graph Convolutional Network

Inspired by two-dimensional graph convolutional networks, Malik *et al.* [45] proposed the M-product based graph convolutional method (TM-GCN) to learn the representation of dynamic graphs [60,61]. For a two-dimensional graph convolutional network with one layer, it can be represented as:

$$Y = \sigma(D^{-0.5}(A+I)D^{-0.5}XW), \tag{5}$$

where $\sigma$ is an activation function, $D$ denotes the degree matrix of $(A + I)$, $X$ and $A$ are the feature matrix and the adjacency matrix respectively. $W$ is the weight parameter that neural network needs to train. Based on two-dimensional graph convolution, the tensor dynamic graph [62,63] embedding is obtained as follow:

$$\overline{\mathbf{X}} = \hat{\sigma}(\mathbf{A} \odot \mathbf{X} \odot \mathbf{W}^{(0)}). \tag{6}$$

The activation function defined as $\hat{\sigma}(\mathbf{X}) = \sigma(\mathbf{X} \times_3 M) \times_3 M^{-1}$, $\mathbf{A}_{::t} = \tilde{A}^{(t)}$ is the frontal slice of adjacency tensor $\mathbf{A} \in \mathbb{R}^{N \times N \times T}$, where $\tilde{A}^{(t)}$ denotes the normalization of $A^{(t)}$, $\mathbf{X}_{::t} = X^{(t)}$ is the frontal slice of feature tensor $\mathbf{X} \in \mathbb{R}^{N \times F \times T}$, $\mathbf{W} \in \mathbb{R}^{F \times F \times T}$ is the weight tensor.

## B. The choice of matrix M

How to choose the $M$ matrix must be considered. We select $M$ as the lower triangular banded matrix for dynamic graphs where $b$ denotes the "bandwidth" of $M$ matrix. This enable each frontal slice $(\mathbf{A} \times_3 M)_{::t}$ to be a linear combination of the adjacency matrices $\mathbf{A}_{::\max(1,t-b+1)}, \ldots, \mathbf{A}_{::t}$. This choice enable each frontal slice exclusively incorporates information from the present and previous slices, in other words, the present front slice aggregates dynamic graph information from the past. For each $t$, $\Sigma_k M_{tk} = 1$, the specific choices are as follows:

$$M_{tk} = \begin{cases} \dfrac{1}{\min(b,t)} & \text{if } \max(1, t-b+1) \leq k \leq t, \\ 0 & \text{otherwise.} \end{cases} \quad (7)$$

## C. Multi-time Segment Modeling

Fig. 3 shows our proposed MS-FTGCN model for traffic flow prediction. It comprises three distinct segments to capture the latent hourly, daily, and weekly dependencies within historical data.

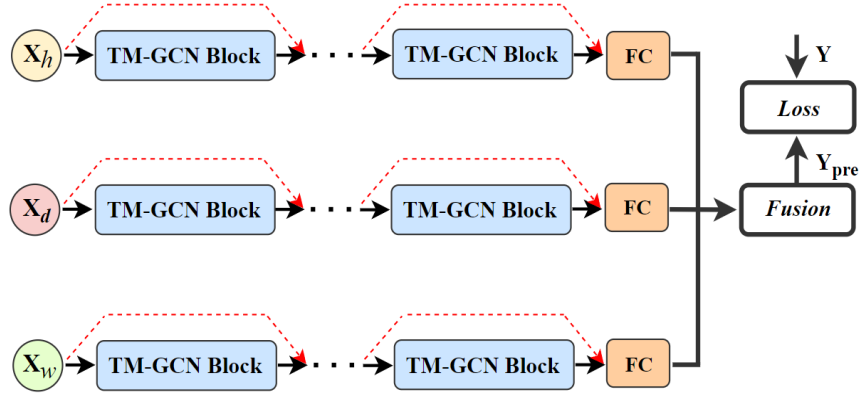

Fig. 3. The framework of MS-FTGCN

Suppose there is a traffic detector that samples at a frequency of $q$ times per day, and assume the current moment is $t_0$. As shown in Fig. 4, In order to predict data with a window size of $T_p$, Three time series slices of length $T_h$, $T_d$, and $T_w$ are taken on the timeline as hourly, daily, and weekly feature inputs. The length of these three time series segments is an integer multiple of the prediction window $T_p$. The following is a detailed description of these three time series segments.

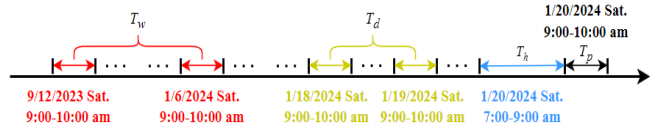

Fig.4. Data extraction of three segments on the timeline

*a)* The *hourly* segment:

$$\mathbf{X}_h = \left(\mathbf{X}_{t_0-T_h}, \mathbf{X}_{t_0-T_h+1}, \ldots, \mathbf{X}_{t_0}\right) \in \mathbb{R}^{N \times N \times T},$$

in the historical time series adjacent to the prediction steps, we consider a segment of data directly linked to the prediction steps, as shown in blue in Fig. 3. It is often the case that the change in traffic flow is not sudden, but rather a gradual process of change. Consequently, the current traffic situation (such as congestion level, traffic flow, etc.) will have an impact on the traffic situation in the next and even future moments. This continuity and interdependence make traffic prediction and planning complex and important.

*b)* The *daily* segment:

$$\mathbf{X}_d = (\mathbf{X}_{t_0-qT_d/T_p+1}, \ldots, \mathbf{X}_{t_0-qT_d/T_p+T_p}, \mathbf{X}_{t_0-q(T_d/T_p-1)+1}, \ldots,$$
$$\mathbf{X}_{t_0-q(T_d/T_p-1)+T_p}, \ldots \mathbf{X}_{t_0-q+1}, \ldots, \mathbf{X}_{t_0-q+T_p}) \in \mathbb{R}^{N \times F \times T_d},$$

as shown in the yellow part in Fig. 4, the daily segment consists of the same time period observed in the past few days as the prediction steps. This is because traffic conditions are cyclical and repeating patterns are very likely to occur, such as morning and evening peak periods on weekdays. We attempt to mine this potential connection from the daily period segment.

*c)* The *weekly* segment:

$$\mathbf{X}_w = (\mathbf{X}_{t_0-7qT_w/T_p+1}, ..., \mathbf{X}_{t_0-7qT_w/T_p+T_p}, \mathbf{X}_{t_0-7q(T_w/T_p-1)+1}, ...,$$
$$\mathbf{X}_{t_0-7q(T_w/T_p-1)+T_p}, ..., \mathbf{X}_{t_0-7q+1}, ..., \mathbf{X}_{t_0-7q+T_p}) \in \mathbb{R}^{F \times N \times T_w},$$

as shown in the red part of Fig. 4, the weekly segment is used to capture periodic information. Generally speaking, Monday's traffic patterns have some similarities with other Monday's traffic patterns in history, but may be significantly different from weekend traffic patterns. Therefore, The weekly cycle segment is aim to capture features associated with the weekly traffic cycle.

*D. Multi-time segment Fusion*

In this section, we will use *Attentional Feature Fusion* (AFF) [64] to fuse three segments. This is a feature fusion mechanism based on multi-scale channel attention modules [65,66] which is shown in Fig. 5. The AFF can be expressed as:

$$\mathbf{Y} = \mathbf{H}(\mathbf{X}_1 \oplus \mathbf{X}_2) \otimes \mathbf{X}_1 + (1 - \mathbf{H}(\mathbf{X}_1 \oplus \mathbf{X}_2)) \otimes \mathbf{X}_2, \tag{8}$$

where **H** is the multi-scale channel attention module, $\otimes$ is the Hadamard product. The red dashed line denotes $1 - \mathbf{H}(\mathbf{X}_1 \oplus \mathbf{X}_2)$, this is due to the fact that the fusion weights $\mathbf{H}(\mathbf{X}_1 \oplus \mathbf{X}_2)$ comprise real number between 0 and 1, similarly, the weights $1 - \mathbf{H}(\mathbf{X}_1 \oplus \mathbf{X}_2)$ also range between 0 and 1, allowing the network to execute weighted averaging or soft selection between $\mathbf{X}_1$ and $\mathbf{X}_2$. It is worth noting that our model has three segments, which means we need to use the AFF twice to fuse the features of the three segments.

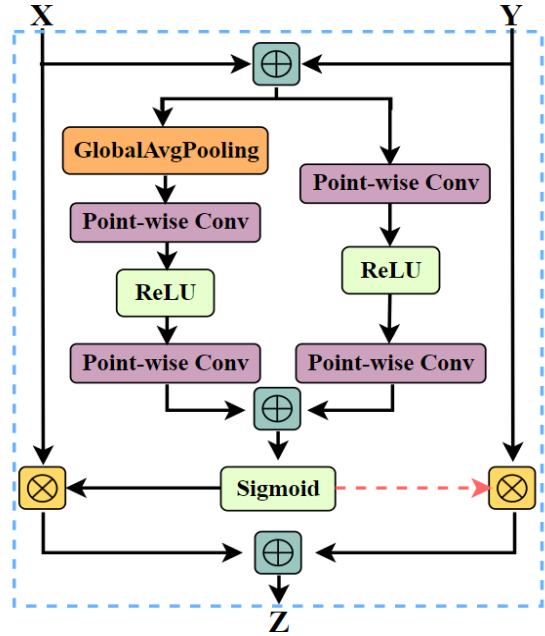

Fig. 5. Attentional Feature Fusion (AFF)

IV. EXPERIMENTS

A series of comparative experiments were conducted on two traffic datasets with the objective of providing a comprehensive evaluation of the effectiveness of the proposed model, offering insights that can inform decision-making in the fields of traffic management and planning.

*A. General Settings*

*Datasets:* PeMS [30,27] offers a unified transportation data repository, collated by California transportation enterprises on highways within the state.

*a) PEMSD8:* PEMSD8 is a dataset consisting of 170 detectors that collect traffic flow data every 5 minutes for a total of 62 days. The dimensions of this dataset are (17856, 170, 3). 170 refers to 170 detectors, 3 refers to each detector collecting data with 3 dimensional features, and 17856 refers to the time slice during this time period. The detector collects data every 5 minutes, so it can collect 12 times within an hour, and there are 24 hours in a day, which takes 62 days.

*b) PEMSD4:* PEMSD4 is a collection of traffic flow data generated over 59 days, with 370 detectors collecting data every 5 minutes. The dimension of this dataset is (16992, 370, 3). These detectors collect data in the same way as the dataset PEMSD8.

It is pertinent to mention that the missing values in the dataset have been substituted with linear interpolation [67,68], then perform a zero mean normalization transformation on the data $x' = x - mean(x)$.

*Evaluation Protocol:* This research employs the Mean Square Error (MSE) between the estimator and the true value as the loss function, with the objective of minimizing this error through backpropagation. The efficacy of our proposed traffic prediction model was evaluated using Mean Absolute Error（MAE）[35,69-75] and Root Mean Square Error (RMSE) [30,76-80] as evaluation metrics. MAE measures the average absolute error between predicted and actual values, while RMSE reflects the root mean square error between predicted and actual values.

$$MAE = \frac{1}{n}\sum_{i=1}^{n}|\hat{y}_i - y_i|, \qquad (9)$$

$$RMSE = \sqrt{\frac{1}{n}\sum_{i=1}^{n}(\hat{y}_i - y_i)^2}, \qquad (10)$$

where $n$ is the total number of samples, $y_i$ and $\hat{y}_i$ are the observed value and predicted value respectively.

*Baselines:* In order to showcase the effectiveness of our proposed model, we conducted comparisons with the following five baseline models: GRU [81-83], FC-LSTM [84,85], ASTGCN [30], GMAN [31], Auto-DSTSGN [86].

Table I Average performance of different methods on PeMSD4 and PeMSD8.

| | PEMSD4 | | | | | | PEMSD8 | | | | | |
|---|---|---|---|---|---|---|---|---|---|---|---|---|
| Horizon | 15min | | 30min | | 60min | | 15min | | 30min | | 60min | |
| | MAE | RMSE | MAE | RMSE | MAE | RMSE | MAE | RMSE | MAE | RMSE | MAE | RMSE |
| FC-LSTM | 24.13 | 38.32 | 25.87 | 40.13 | 27.25 | 41.52 | 20.73 | 31.66 | 21.44 | 32.87 | 22.32 | 34.13 |
| GRU | 18.23 | 29.08 | 19.88 | 30.58 | 21.45 | 32.07 | 17.25 | 25.32 | 18.12 | 26.89 | 19.21 | 28.44 |
| ASTGCN | 19.27 | 31.25 | 19.92 | 32.43 | 22.31 | 34.13 | 16.08 | 24.24 | 16.72 | 25.52 | 17.56 | 27.30 |
| GMAN | 14.95 | 28.25 | 16.07 | 29.78 | 17.58 | 31.52 | 14.56 | 22.45 | 15.29 | 24.03 | 15.74 | 26.52 |
| Auto-DSTSGN | 14.97 | 28.35 | 16.31 | 29.73 | 18.15 | 31.44 | 13.98 | 22.29 | 14.39 | 23.54 | 16.14 | 25.78 |
| **MS-FTGCN(ours)** | **14.69** | **28.13** | **15.87** | **29.38** | **17.48** | **30.72** | **13.77** | **21.84** | **14.05** | **23.18** | **15.69** | **25.17** |

## B. Parameter Analysis

As explained in the previous section, selecting the appropriate bandwidth for the *M* matrix ensures that each frontal slice only includes information from past and current slices that are temporally close. It is important to note that different bandwidths can have varying impacts on experimental results. This section will use the prediction of 60 minutes as an example to examine the effects of different *M* matrix bandwidths on experimental outcomes. Fig. 6 shows that the bandwidth selection range is from 1 to 12. This is because the original dataset is sampled every 5 minutes, and the 60 minutes prediction sample is composed of 12 time slices.

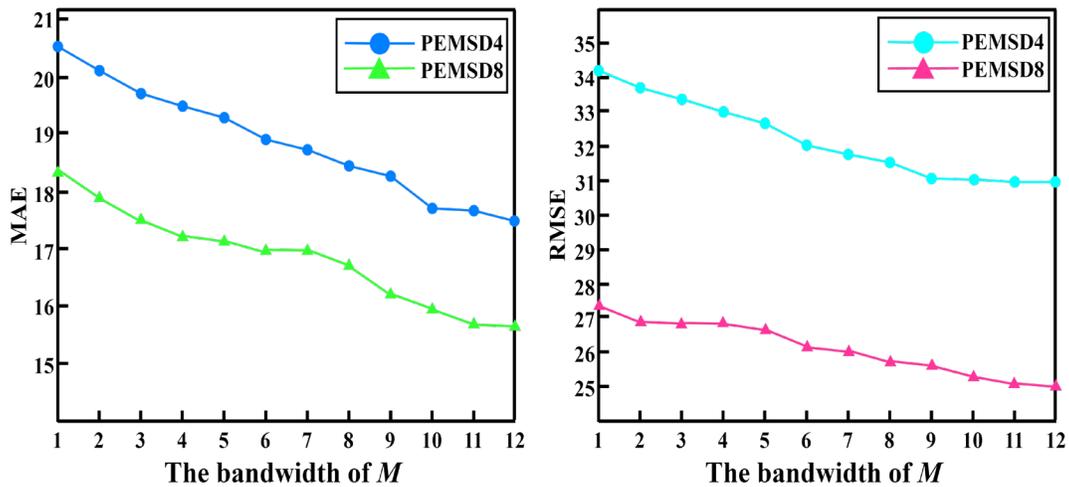

Fig. 6. The impact of different bandwidth of *M* on dataset PEMSD4 and PEMSD8

Fig. 6 illustrates that augmenting the bandwidth of *M* results in a progressive enhancement of prediction performance. This phenomenon can be elucidated as the quantity of information amassed in each frontal slice increases in tandem with the expansion of bandwidth. This signifies that a more extensive *M* matrix can capture a greater number of time series correlations, thereby fortifying the capacity for precise prediction of traffic flow. Furthermore, as bandwidth expands, the model can more effectively leverage potential patterns and trends in time series data, further optimizing its performance. In prediction of traffic flow for 15/30/60 minutes, MAE and RMSE reached their lowest.

*C. Comparison Performance*

Here, in the experiment, we compared the traffic prediction performance of different methods in 15/30/60 minutes. Table II shows the comparison results of each model on MAE/RMSE. The MS-FTGCN model we proposed has remarkable performance in traffic flow prediction compared to other models. The analysis of the experimental results allows us to draw several important conclusions:

*a)* The benefits of using neural networks to model nonlinear traffic flow data have been demonstrated. Deep learning methods generally outperform traditional machine learning techniques in capturing intricate data patterns and dynamic changes. This is due to the strong non-linear fitting ability of neural networks, which can automatically learn features and relationships without the need for manual feature engineering.

*b)* Compared to other methods, MS-FTGCN has the best predictive performance and shows more obvious advantages in long-term prediction. This is because the M-transform in TM-GCN aggregates feature information from different historical moments based on different lengths of bandwidth, and then considers the various impacts generated by sensor network nodes at different historical moments. Therefore, it can capture spatial-temporal features more effectively.

## V. CONCLUSION

This paper introduces a spatial-temporal graph convolutional model named MS-FTGCN, which has been indicated to be effective in traffic flow prediction. Evaluation on two real-world datasets indicates the model outperforms baseline methods in terms of predictive accuracy.

In the next, we will take into account the impact of weather factors on traffic flow and conduct comparative experiments to get more comprehensive prediction results. Additionally, we will examine the influence of various configurations of the *M* matrix on prediction outcomes, aiming to iteratively enhance both the efficiency and accuracy.